\documentclass[conference]{IEEEtran}
\IEEEoverridecommandlockouts
\usepackage{cite}
\usepackage{amsmath,amssymb,amsfonts}
\usepackage{algorithmic}
\usepackage{graphicx}
\usepackage{textcomp}
\usepackage{xcolor}
\usepackage{amsmath,epsfig}
\usepackage{bbm}
\usepackage[T1]{fontenc}
\usepackage{array}
\usepackage{booktabs}
\usepackage{comment}
\usepackage{makecell}
\usepackage{graphicx,subfigure}
\usepackage{hyperref}
\def\BibTeX{{\rm B\kern-.05em{\sc i\kern-.025em b}\kern-.08em
    T\kern-.1667em\lower.7ex\hbox{E}\kern-.125emX}}
\begin{document}

\title{Personalized Food Image Classification: \\ Benchmark Datasets and New Baseline}

\author{\IEEEauthorblockN{Xinyue Pan,
Jiangpeng He, and
Fengqing Zhu}
\IEEEauthorblockA{Elmore Family School of Electrical and Computer Engineering, Purdue University, West Lafayette, IN 47906, U.S.A.}}
\maketitle

\begin{abstract}


    

    
    
    

Food image classification is a fundamental step of image-based dietary assessment, enabling automated nutrient analysis from food images. Many current methods employ deep neural networks to train on generic food image datasets that do not reflect the dynamism of real-life food consumption patterns, in which food images appear sequentially over time, reflecting the progression of what an individual consumes. Personalized food classification aims to address this problem by training a deep neural network using food images that reflect the consumption pattern of each individual. However, this problem is under-explored and there is a lack of benchmark datasets with individualized food consumption patterns due to the difficulty in data collection. In this work, we first introduce two benchmark personalized datasets including the Food101-Personal, which is created based on surveys of daily dietary patterns from participants in the real world, and the VFN-Personal, which is developed based on a dietary study. In addition, we propose a new framework for personalized food image classification by leveraging self-supervised learning and temporal image feature information. Our method is evaluated on both benchmark datasets and shows improved performance compared to existing works. The dataset has been made available at: \url{https://skynet.ecn.purdue.edu/~pan161/dataset_personal.html}

\end{abstract}

\begin{IEEEkeywords}
Food image classification, personalized classifier, image-based dietary assessment, self-supervised learning
\end{IEEEkeywords}

\vspace{-1mm}
\section{Introduction}

\label{sec:intro}
Food image classification is crucial for image-based dietary assessment, which aims to provide an accurate profile of foods consumed and their portion sizes based on an individual's habitual dietary intake \cite{boushey2017}. Given the widespread use of mobile devices, many individuals now utilize food logging apps to daily track their food intake, aiding in maintaining a healthy diet over time \cite{SAMAD20222, Coughlin2015}.

\begin{figure}[t]
\centering
\includegraphics[width=1\linewidth]{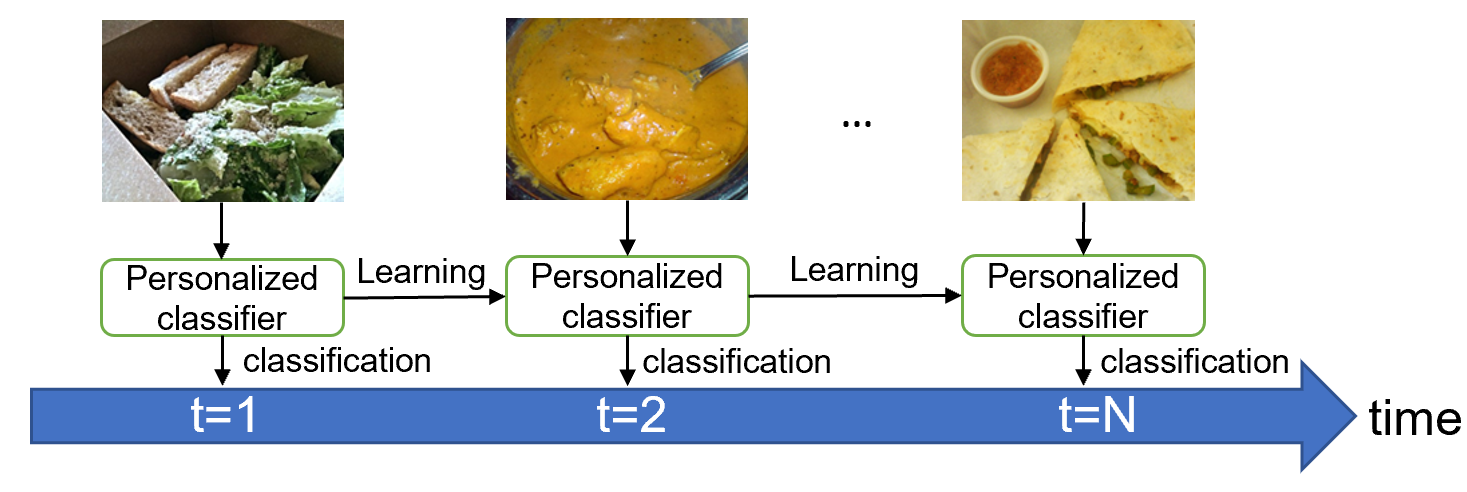}
\caption{An illustration of personalized food image classification. The objective is to train a personalized classifier based on food consumption patterns to improve food classification performance. }
\label{fig:pipeline}
\vspace{-5mm} 
\end{figure}

Although existing works \cite{Mao2021ImprovingDA, mao2021, he2020_mipr, he2021end, he2022long, pan2023muti, he2023singlestage} have demonstrated promising results using static food datasets, food image classification is much more challenging in real-world settings where data comes sequentially overtime~\cite{raghavan2023online, he2021_iccvw, ILIO, he2022, he2023longtailed, he2022exemplar}. The most recent work focuses on addressing this issue for each individual by designing a personalized food classifier~\cite{8451422, Horiguchi2018PersonalizedCF}. In such contexts, individuals capture food images in sequence, thereby documenting their dietary habits chronologically. We refer to this sequential data as a "food consumption pattern". A Food consumption pattern typically exhibits unbalanced food distribution, diverse cooking styles, and previously unseen food classes over time \cite{Kitamura2010}. The main objective of personalized food classification is to classify each food image as it appears sequentially over time in a food consumption pattern. This ensures enhanced classification accuracy tailored to a person's unique dietary progression. Fig. \ref{fig:pipeline} shows an illustration of personalized food image classification, which learns the food class appeared in a food consumption pattern over time. However, there exist two major challenges. The first is a lack of publicly available benchmark personalized food image datasets. This is mainly due to the difficulty in collecting food consumption patterns from different individuals over time.  The second is a lack of exploration into learning sequential image data streams containing previously unseen food classes and associated contextual information from the food consumption pattern. 

Our work aims to address both aforementioned challenges by creating benchmark datasets encapsulating personalized food consumption patterns and developing a novel personalized classifier to improve the performance of existing methods \cite{1053964, Horiguchi2018PersonalizedCF,8451422}. To address the first challenge of lacking available datasets, we first introduce two benchmark personalized food consumption datasets by leveraging two public food image datasets \cite{bossard14,mao2021} with food categories and proceed as follows. For both datasets, we have short-term food consumption patterns from volunteers' input. We then extend and simulate different patterns using a method based on \cite{pan2022}.

Existing personalized food classification methods \cite{Horiguchi2018PersonalizedCF,8451422} store food image features extracted from pre-trained models and employ the nearest-class-mean classifier and nearest neighbor classifier to adapt each individual's eating habit. The most recent work \cite{kim2021} further improves the performance by sharing food records across multiple food consumption patterns. Nonetheless, these approaches exhibit several limitations. Firstly, while processing each image in a consumption pattern, the pre-trained feature extractor remains static, unable to learn and update dynamically using new food images. Secondly, existing work only considers the short-term frequency of food occurrence, lacking the exploration into temporal contextual information, which provides diet change over the time.

In this work, we introduce a personalized classifier that addresses all the aforementioned limitations. By enhancing the image feature extraction with self-supervised learning, our model updates dynamically with each new food image. Moreover, we enrich the temporal context by concatenating image features within a sliding window, facilitating a deeper consideration of image feature-based temporal nuances. The main contributions of our work can be summarized as follows:
\vspace{-0.5mm}
\begin{itemize}
    \item We introduce two new benchmark datasets for personalized food image classification including the \textbf{Food101-Personal} and the \textbf{VFN-Personal}, and we have made it open to the public.
    \item We propose a novel personalized classifier through feature extraction update using self-supervised learning and a sliding window technique to capture temporal contextual information based on image features.
\end{itemize}

\vspace{-1mm}
\section{Benchmark Datasets}
\label{sec:data}
In this section, we introduce two benchmark personalized datasets including Food101-Personal and VFN-Personal. Unlike existing food image classification methods that are trained on public food datasets~\cite{bossard14, mao2021, Min-ISIA-500-MM2020}, there is no publicly available personalized food image dataset due to the challenges in obtaining food consumption patterns for each individual, which reflects their dietary habits over time. Our work addresses this gap by first collecting short-term food consumption patterns through surveys or dietary studies and then simulating the long-term personalized food consumption patterns following the method in \cite{pan2022} where a modified Markov chain is used to capture temporal contextual information based on the provided initial short-term food consumption pattern.

\noindent\textbf{Food101-Personal: }We conducted an online survey using the Food-101 dataset \cite{bossard14}, where participants were asked to simulate one week of food consumption patterns by selecting foods from the 101 classes in Food-101. We collected 20 participants' patterns, each with over 20 food records, and simulated long-term patterns using the method described in \cite{pan2022}. To develop a more representative benchmark, we cluster food images within each food class from the Food-101 dataset and employ a Gaussian distribution model as described in \cite{pan2022} to create intra-class dissimilarities within each class in a pattern. Overall, the benchmark includes 20 patterns with 300 images each and an average of 44 food classes per pattern. 

\noindent\textbf{VFN-Personal: }For the VFN dataset \cite{mao2021}, we conducted a dietary study from healthy participants aged 18 to 65 using the image-based dietary assessment system~\cite{shao2021}. Participants captured images of foods they consumed for three days. We collected data from over 70 participants, retaining 26 short-term patterns which have at least 15 records each. Similar to the Food101-Personal dataset, we employed the method in \cite{pan2022} to simulate long-term food consumption patterns. Overall, the VFN-Personal dataset comprises 26 patterns, each containing 300 images and an average of 29 food classes per pattern.

\vspace{-1mm}
\section{Method}
\label{sec:method}
In this section, we introduce a novel method to improve the accuracy of personalized food image classification. Our approach consists of two key components: (1) employing self-supervised learning to update the feature extractor, as described in Section \ref{sec:mod_up}, and (2) using a sliding window to capture multiple-image temporal information within a food consumption pattern, as explained in Section \ref{sec:slide}.

\vspace{-1mm}
\subsection{Feature Extraction Using Self-supervised Learning}
\label{sec:mod_up}




One limitation of existing personalized food classification approaches is the fixed feature extractor, which is unable to update using new images in a food consumption pattern. In this paper, we address these issues by leveraging self-supervised learning~\cite{chen2020,jure2021,chen2020simsiam, peng2023self} to learn image features without ground truth labels. Our method is designed to be compatible with any self-supervised learning backbone.

To accommodate self-supervised learning in our scenario where a large training batch is not feasible as new images typically arrive sequentially one by one, we apply the following techniques to create representative input batches.

\noindent\textbf{Group normalization:} In existing self-supervised learning with batch normalization, the error tends to increase rapidly as the batch size decreases. However, utilizing large batch sizes in the early time steps is not feasible in our scenario due to the limited number of food images. To tackle this issue, we replace batch normalization layers with group normalization layers~\cite{wu2018}, which provides constant error across different batch sizes, making it a more reliable alternative.

\noindent\textbf{Random image sampling}\label{sec:rs} 
We employ a random image sampling technique as described in \cite{ValizadeganAH12} to select input images for the self-supervised learning algorithm. Let $t$ denote the current time step. Our objective is to randomly sample images from time steps before $t$, rather than sampling them in a consecutive temporal order. The input set of images can be denoted as $I_t = [f_a, f_b, \dots], \; 1 \le a,b,\dots \le t$, where $a, b, \dots$ represent the sampling time steps.




\noindent\textbf{Dual Instances Learning}\label{sec:dil} To tackle the issue of class imbalance and intra-class variability in food classification, we propose to use a pair of images ($f_{i,1}$, $f_{i,2}$) from each class $i$ rather than employing two augmentations of the same image as inputs. The motivation is that different images from the same class should exhibit similar feature representation.
 
\vspace{-1mm}
\subsection{Sliding Window}
\label{sec:slide}

Existing personalized food classification methods\cite{Horiguchi2018PersonalizedCF,8451422} rely on a single image feature to classify images within individual consumption patterns. 
However, incorporating temporal contextual information based on multiple images within food consumption patterns is also important to help capture the unique diet characteristics of each individual. In this work, we propose to combine the single-image feature and multiple-image temporal information for classification where the latter is achieved by constructing the sliding window to capture past multiple-image temporal information based on concatenated image features.

\begin{figure}[t]
    \centering
    \includegraphics[width=0.99\linewidth]{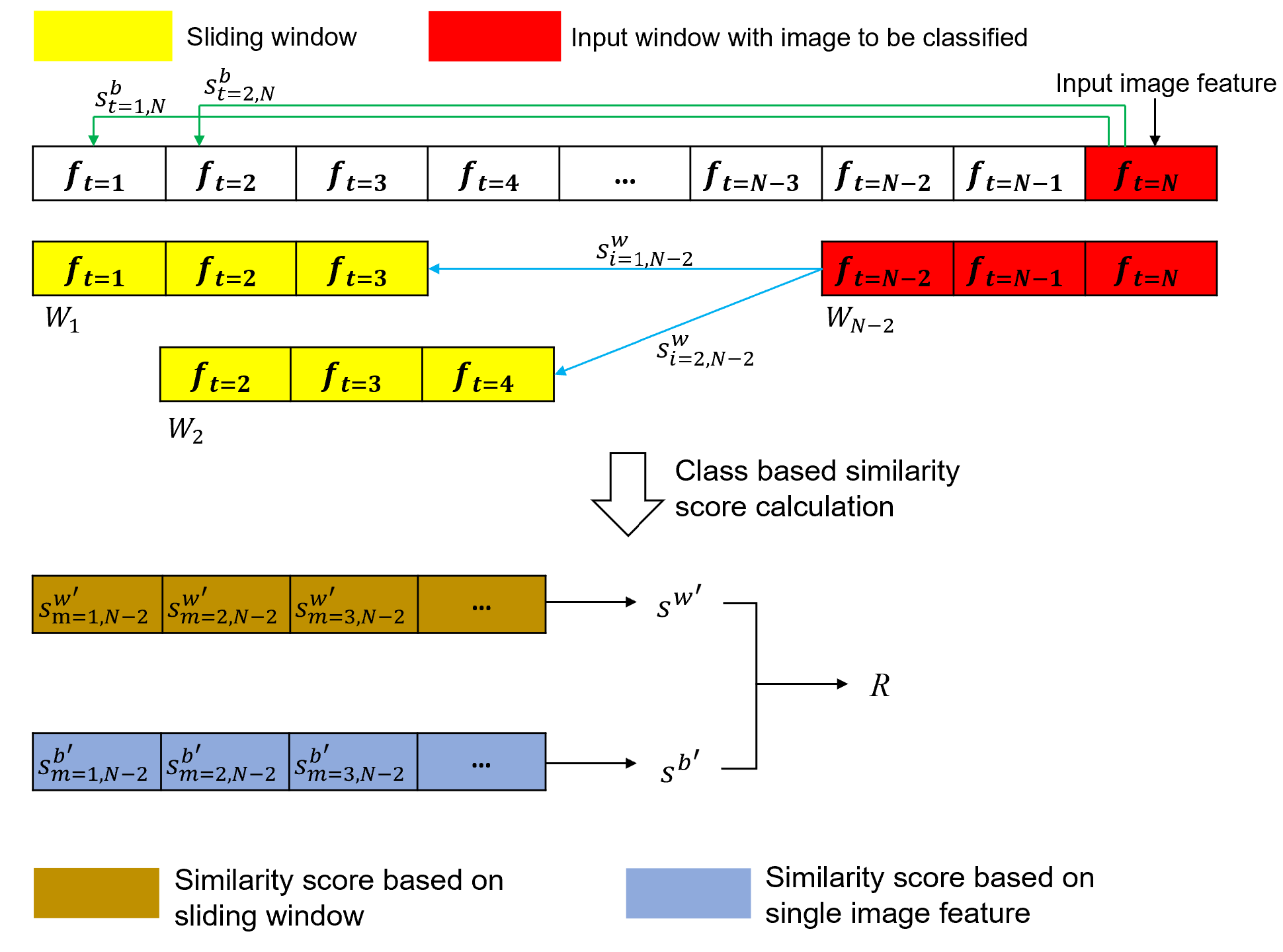}
    \caption{Overview of the sliding window method. We obtain the similarity score for each food class by using both the current image feature and sliding windows, denoted as $s^{b}$ and $s^{w}$, respectively. The final prediction is computed based on the combined similarity vector $R$. }
    \label{fig:slide2}
\vspace{-2mm}
\end{figure}

Specifically, we first compute single-image similarity score  $s^b_{t,N}$, to find the image that is most similar to the image to be classified. Given a new image at $t=N$ with number of $M$ food classes appeared so far, we calculate $s^b_{t,N}$ by finding the cosine similarity between input image feature $f_N$ and previous image features $f_t, t \in {1,2,...N-1}$, with the formula of 
\begin{equation}
\begin{split}
    s^b_{t,N} = \frac{f^{T}_{t}f_{N}}{||f_{t}||_2||f_{N}||_2}, \; 1 \le t < N\;\;\;\; 
\end{split}
\end{equation}
where $T$ corresponds to the transpose of a matrix. Since the same food class may appear multiple times before $t = N$, we first take the maximum similarity among image features in the same class before $t=N$, 
denoted as $s^b_{m},1\le m\le M$
and then apply softmax to get $s^{b'}_{m}$.
where $m$ denotes the food class index and $c_t$ denotes the food class at time step $t$.

Each sliding window $W$ can be built by concatenating image features as represented in the follows:
\begin{equation}
    W_i = ([f_{i},f_{i+1},...,f_{i+k-1}],c_{i+k-1}), \; 1 \le i \le N-k+1
\end{equation}
where $i$ denotes the sliding window index, $k$ is the length of the window, and $c_{i+k-1}$ represents the class label associated with the window $W_i$.
To find the $W_i$ with the highest similarity to $W_{N-k+1}$, which contains the current image to be classified, we apply the nearest neighbor classifier to calculate the cosine similarity among sliding windows as

\begin{equation}
   s^w_{i,N-k+1}=\frac{W^T_{i} W_{N-k+1}}{||W_{i}||_2||W_{N-k+1}||_2}
\end{equation}

For the food class-based similarity score, we take the maximum value among all windows belonging to the same food label, denoted as $s^{w}_{m}, 1 \le m \le M$.  
We then take the softmax of $s^w_{m}$ and denote it as $s^{w'}_{m}$.

Finally, we combine the similarity scores from $s^{w'}$ and $s^{b'}$ to obtain $R_m$, which is computed as follows: $R_m = s^{b'}_{m} (s^{w'}_{m})^{\alpha}, \; 0 \le \alpha \le 1$
where $\alpha$ denotes the weight associated with $s^{w'}$, which controls the level of significance of the sliding window method in computing the final similarity score. The higher the $\alpha$ value, the greater the level of significance. The final prediction is calculated by: $p_t = {argmax}\{R_m\}$, where $t$ denotes the time step at which the image is to be classified.

\begin{table*}[t]
\begin{center}
\caption{Results of personalized food classification on two benchmark datasets. Each cell represents the top-1 $classification\; accuracy \pm std$ across different patterns at different time steps. } \label{tab:result1}
\begin{tabular}{c|c|c|c|c||c|c|c|c}
  \toprule
  &\multicolumn{4}{c}{Food101-Personal} &\multicolumn{4}{c}{VFN-Personal} \\
  Method & $t_{75}$ & $t_{150}$ & $t_{225}$ &$t_{300}$ & $t_{75}$ & $t_{150}$ & $t_{225}$ &$t_{300}$ 
  \\
  \bottomrule
  & \multicolumn{8}{c}{Comparison with existing work} \\
  \toprule 
  \textbf{CNN} \cite{Dai2021} &    16.7{\footnotesize $\pm7.0$}&
  16.3{\footnotesize $\pm6.3$}&
  16.2{\footnotesize $\pm6.5$}& 
  16.6{\footnotesize $\pm6.9$}& 
  3.1{\footnotesize $\pm3.8$}& 3.2{\footnotesize $\pm3.4$}&
  3.1{\footnotesize $\pm2.6$}&
  3.6{\footnotesize $\pm3.2$}\\
   \textbf{SVMIL}\cite{liu2004} &   23.4{\footnotesize $\pm12.6$}&
  34.1{\footnotesize $\pm11.7$}&
  40.2{\footnotesize $\pm10.9$}& 
  44.3{\footnotesize $\pm9.8$}& 
  42.0{\footnotesize $\pm10.2$}& 51.6{\footnotesize $\pm10.4$}&
  56.1{\footnotesize $\pm10.3$}&
  58.8{\footnotesize $\pm10.1$}\\
  \textbf{1-NN} \cite{1053964} &    
  31.6{\footnotesize $\pm14.3$}&
  42.8{\footnotesize $\pm12.8$}&
  48.8{\footnotesize $\pm11.7$}& 
  53.0{\footnotesize $\pm10.5$}& 
  53.4{\footnotesize $\pm9.0$}& 
  63.2{\footnotesize $\pm9.2$}&
  67.3{\footnotesize $\pm9.2$}&
  69.4{\footnotesize $\pm8.9$}\\
    \textbf{SPC} \cite{Horiguchi2018PersonalizedCF}  & 39.1{\footnotesize $\pm13.6$}& 51.6{\footnotesize $\pm12.1$} & 57.7{\footnotesize $\pm10.4$} & 
    61.6{\footnotesize $\pm9.3$}& 
    56.7{\footnotesize $\pm7.7$}& 66.5{\footnotesize $\pm7.6$}&  70.6{\footnotesize $\pm7.6$} &  73.0{\footnotesize $\pm7.1$}\\
    \textbf{SPC++} \cite{8451422} & 39.0{\footnotesize $\pm13.6$} & 51.4{\footnotesize $\pm11.8$} & 57.6{\footnotesize $\pm10.8$}&
    61.7{\footnotesize $\pm9.1$}& 
    57.0{\footnotesize $\pm7.4$}& 67.0{\footnotesize $\pm7.2$} &  70.9{\footnotesize $\pm7.3$} &  73.2{\footnotesize $\pm6.6$}\\
    \textbf{Ours with SimSiam}& \textbf{40.1}{\footnotesize $\pm13.3$} & 53.4{\footnotesize $\pm11.2$} & 59.8{\footnotesize $\pm9.6$}& 64.0{\footnotesize $\pm8.5$} & \textbf{57.6}{\footnotesize $\pm7.8$}  & 67.8{\footnotesize $\pm7.6$} & 71.8{\footnotesize $\pm7.6$}& \textbf{74.1}{\footnotesize $\pm7.0$}\\
    \textbf{Ours with Barlow Twins}& \textbf{40.1}{\footnotesize $\pm13.2$} & \textbf{53.8}{\footnotesize $\pm10.8$} & \textbf{60.2}{\footnotesize $\pm9.4$}& \textbf{64.1}{\footnotesize $\pm8.3$}  & 57.4{\footnotesize $\pm7.9$}  & \textbf{67.9}{\footnotesize $\pm7.4$} & \textbf{71.9}{\footnotesize $\pm7.6$}& 74.0{\footnotesize $\pm7.2$}\\
  \bottomrule
  & \multicolumn{8}{c}{Ablation studies with SimSiam as backbone} \\
  \toprule
    \textbf{RS+SPC\textsl{++}} & 39.1{\footnotesize $\pm13.8$} & 51.7{\footnotesize $\pm11.7$} & 58.0{\footnotesize $\pm10.3$} & 
    62.1{\footnotesize $\pm9.1$}& 
    57.2{\footnotesize $\pm7.9$}& 66.9{\footnotesize $\pm7.7$}  & 71.2{\footnotesize $\pm7.7$} & 73.6{\footnotesize $\pm7.3$}\\
    \textbf{DIL+SPC\textsl{++}} & 39.5{\footnotesize $\pm13.8$} & 51.9{\footnotesize $\pm12.0$} & 58.0{\footnotesize $\pm10.3$} & 
    62.1{\footnotesize $\pm9.1$}& 
    \textbf{58.0}{\footnotesize $\pm7.5$}& 67.8{\footnotesize $\pm7.3$}  & 71.5{\footnotesize $\pm7.4$} & 74.0{\footnotesize $\pm6.7$}\\
    \textbf{RS+DIL+SPC\textsl{++}} & 39.7{\footnotesize $\pm14.3$} & 52.3{\footnotesize $\pm12.0$} & 58.4{\footnotesize $\pm10.3$}& 62.4{\footnotesize $\pm9.3$} & 57.4{\footnotesize $\pm8.3$} &
    67.5{\footnotesize $\pm7.3$}& 
    71.5{\footnotesize $\pm7.8$}& 73.9{\footnotesize $\pm7.2$}\\
    \textbf{RS+SW}& 40.0{\footnotesize $\pm13.2$} & \textbf{53.6}{\footnotesize $\pm11.3$} & 59.9{\footnotesize $\pm9.9$}& 64.0{\footnotesize $\pm8.9$} & 57.6{\footnotesize $\pm8.1$} &
    67.8{\footnotesize $\pm7.8$}& 
    71.8{\footnotesize $\pm7.8$}& 73.9{\footnotesize $\pm7.2$}\\
    \textbf{DIL+SW}& 40.0{\footnotesize $\pm13.2$} & 53.5{\footnotesize $\pm11.5$} & \textbf{59.9}{\footnotesize $\pm9.9$}& 63.9{\footnotesize $\pm8.8$} & 57.5{\footnotesize $\pm8.0$} &
    67.7{\footnotesize $\pm7.7$}& 
    71.8{\footnotesize $\pm7.7$}& 73.8{\footnotesize $\pm7.2$}\\
    \textbf{RS+DIL+SW(Ours)}& \textbf{40.1}{\footnotesize $\pm13.3$} & 53.4{\footnotesize $\pm11.2$} & 59.8{\footnotesize $\pm9.6$}& \textbf{64.0}{\footnotesize $\pm8.5$} & 57.6{\footnotesize $\pm7.8$}  & \textbf{67.8}{\footnotesize $\pm7.6$} & \textbf{71.8}{\footnotesize $\pm7.6$}& \textbf{74.1}{\footnotesize $\pm7.0$}\\
  \bottomrule
  & \multicolumn{8}{c}{Ablation studies with Barlow Twins as backbone} \\
  \toprule 
   \textbf{RS+SPC\textsl{++}} &  39.8{\footnotesize $\pm13.5$}  & 51.8{\footnotesize $\pm11.0$} &  57.7{\footnotesize $\pm10.1$}& 61.7{\footnotesize $\pm8.8$} &
   57.7{\footnotesize $\pm7.4$}& 
   67.4{\footnotesize $\pm7.5$}& 70.9{\footnotesize $\pm7.8$} &  73.0{\footnotesize $\pm7.0$}\\
   \textbf{DIL+SPC\textsl{++}} &  39.5{\footnotesize $\pm13.6$}  & 52.1{\footnotesize $\pm11.2$} &  58.4{\footnotesize $\pm9.6$}& 62.4{\footnotesize $\pm8.4$} &
   \textbf{57.9}{\footnotesize $\pm7.4$}& 
   67.8{\footnotesize $\pm7.3$}& 71.5{\footnotesize $\pm7.1$} &  73.6{\footnotesize $\pm6.7$}\\
    \textbf{RS+DIL+SPC\textsl{++}} & 39.6{\footnotesize $\pm14.0$} &  52.1{\footnotesize $\pm11.4$} & 58.6{\footnotesize $\pm9.7$}&  62.6{\footnotesize $\pm8.4$} & 57.8{\footnotesize $\pm7.4$} &  67.3{\footnotesize $\pm7.4$} &  71.3{\footnotesize $\pm7.4$} &  73.7{\footnotesize $\pm7.1$}\\
    \textbf{RS+SW}& \textbf{40.3}{\footnotesize $\pm12.6$} & 53.7{\footnotesize $\pm10.7$} & 59.6{\footnotesize $\pm9.7$}& 63.4{\footnotesize $\pm8.6$} & 57.7{\footnotesize $\pm8.0$} &
    67.8{\footnotesize $\pm7.7$}& 
    71.8{\footnotesize $\pm7.7$}& 74.0{\footnotesize $\pm7.1$}\\
    \textbf{DIL+SW} & 40.2{\footnotesize $\pm13.3$} &  53.7{\footnotesize $\pm10.9$} & 60.0{\footnotesize $\pm9.7$}&  63.9{\footnotesize $\pm8.6$} & 57.7{\footnotesize $\pm7.8$} &  67.6{\footnotesize $\pm7.5$} &  71.7{\footnotesize $\pm7.5$} &  73.7{\footnotesize $\pm7.3$}\\
    \textbf{RS+DIL+SW(Ours)}& 40.1{\footnotesize $\pm13.2$} & \textbf{53.8}{\footnotesize $\pm10.8$} & \textbf{60.2}{\footnotesize $\pm9.4$}& \textbf{64.1}{\footnotesize $\pm8.3$}  & 57.4{\footnotesize $\pm7.9$}  & \textbf{67.9}{\footnotesize $\pm7.4$} & \textbf{71.9}{\footnotesize $\pm7.6$}& \textbf{74.0}{\footnotesize $\pm7.2$}\\ 
\bottomrule
\end{tabular}
\end{center}
\end{table*}
\vspace{-0.5mm}

\section{Experiments}
\label{sec:exp}
In this section, we evaluate our proposed methods by comparing with existing works on Food101-Personal and VFN-Personal datasets introduced in Section~\ref{sec:data}. We also conduct an ablation study to demonstrate the effectiveness of each component in our proposed framework. 

\subsection{Benchmark Protocol}
Different from the general image classification task that train a model on training data and evaluate on test data, there is no split of train and test in personalized dataset. Therefore, we propose the following evaluation protocol. Given a personalized dataset containing multiple food consumption patterns from different individuals, the personalized classifier is evaluated on each pattern one by one by assuming (1) the data becomes available sequentially, and (2) the model is updated in an online scenario, \textit{i.e.}, the training epoch is 1. During each time step in a pattern, the model first make prediction on the new image as one of the food classes seen so far and then use it for update. The performance on each pattern is evaluated by calculating the cumulative mean accuracy at each time step as:
\begin{equation}
    C\_accuracy(t) = \frac{1}{t}\Sigma_{\tau = 1} ^ {\tau=t} \mathbbm{1}(p_{\tau} = c_{\tau})
\end{equation}
$\mathbbm{1}(\cdot)$ is a function indicating whether the current prediction of food is correct or not. $p_{\tau}$ denotes the prediction at time step $\tau$ for a pattern, and $c_{\tau}$ represents the class label at time step $\tau$ for a pattern. The overall performance on the entire dataset is calculated as the mean accuracy for all the personalized food consumption patterns.

\subsection{Experiment Setup}
\noindent\textbf{Methods for comparison: }We employ Simsiam \cite{chen2020simsiam} and Barlow Twins \cite{jure2021} as self-supervised learning backbones, replacing batch normalization layers with group normalization layers \cite{wu2018} for small input batch sizes. We compare our method with existing methods including \textbf{CNN} \cite{Dai2021}, which uses a general fixed-class convolutional neural network on ISIA-500 dataset \cite{Min-ISIA-500-MM2020}; \textbf{1-NN} \cite{1053964}, which is a one nearest neighbor method; \textbf{SVMIL} \cite{liu2004}, a common incremental learning method that utilizes the SVM model and updates the model based on a new image feature at every single time step; \textbf{SPC} \cite{Horiguchi2018PersonalizedCF} and \textbf{SPC\textsl{++}} \cite{8451422}, which employs nearest neighbor and nearest class mean classifier with a fixed pre-trained model, and incorporating a time-dependent model and weight optimization for classification. 

Furthermore, we conduct an ablation study to demonstrate the effectiveness of each proposed component including \textbf{Random Sampling (RS)}, which is a random sampling method illustrated in section \ref{sec:rs}; \textbf{Dual Instance Learning (DIL)}, which is the dual instance learning approach described in section \ref{sec:dil}; and \textbf{Sliding Window (SW)}, which employs the sliding window method to capture multiple-image temporal information, as explained in Section \ref{sec:slide}.

\begin{figure}[t]
\centering
\includegraphics[width=1.0\linewidth]{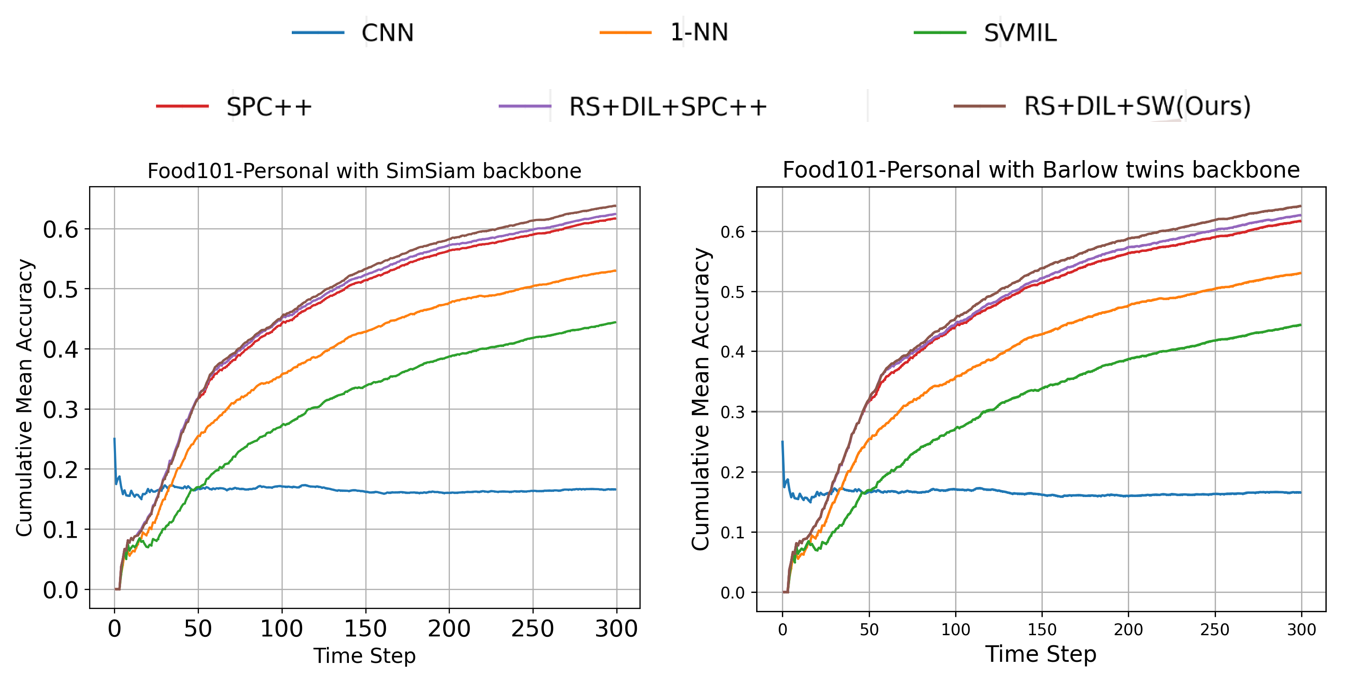}
\caption{Classification accuracy at each time stamp on Food101-Personal datasets}
\label{fig:results}
\end{figure}

\noindent\textbf{Implementation detail:} We utilize ResNet-50~\cite{he2015} pre-trained on ISIA-500 dataset~\cite{Min-ISIA-500-MM2020} as backbone to extract image features in food consumption patterns. The batch size is set to 32 with training epochs 1 in online scenario. For SimSiam, we use SGD optimizer with learning rate of 0.001 and weight decay 0.0001. For Barlow Twins, we utilize the LARS optimizer \cite{yang2017} with a learning rate of $1 \times 10^{-6}$ on normalization and bias layers and $1 \times 10^{-5}$ on other layers, along with a weight decay of $1 \times 10^{-6}$. For our sliding window \textbf{SW}, we empirically set $\alpha = 0.0025$ with a window size of $k = 5$. The method is applied when $t \ge 50$.

\subsection{Results and Discussion}
Table \ref{tab:result1} shows the results of personalized food classification at selected time steps for the Food101-Personal and VFN-Personal dataset. The first part of table \ref{tab:result1} shows the comparison of classification performance of our proposed method with existing works. In the existing works, \textbf{CNN} \cite{Dai2021} constantly exhibits low accuracy over time, as it does not learn to classify new classes from the consumption patterns. \textbf{SVMIL} underperforms compared to \textbf{1-NN}, due to only having one new image at each time step to learn from and not addressing the mini-batch learning issue. \textbf{1-NN}\cite{1053964} shows inferior performance compared to \textbf{SPC} \cite{Horiguchi2018PersonalizedCF} because of not considering cold start problem. 
\textbf{SPC\textsl{++}} \cite{8451422} outperforms \textbf{SPC} \cite{Horiguchi2018PersonalizedCF} by taking into account the short-term frequency of food consumption. Our proposed method can outperform the existing works at most time steps by considering the image feature updates during training in food consumption patterns over time and multiple-image temporal information. Our method improves the classification accuracy for $2.6\%$ and $1\%$ on \textbf{Food101-Personal} and \textbf{VFN-Personal} dataset, respectively. 

The second and third part of table \ref{tab:result1} shows ablation studies of our proposed method with SimSiam and Barlow Twins as backbone respectively.
From \textbf{RS+SPC\textsl{++}} and \textbf{DIL+SPC\textsl{++}} method, it can be observed that both \textbf{RS} and \textbf{DIL} contribute nearly equally to the improvement of classification accuracy, indicating their equal effectiveness in sampling input images for self-supervised learning. Integrating both methods (i.e. \textbf{RS+DIL+SPC\textsl{++}}) leads to further improvement of classification accuracy since it facilitates learning from a balanced class distribution, considers intra-class dissimilarity within a class and general image features without memorizing the specific appearance order of images within a pattern. Moreover, integrating one of the sampling techniques with the \textbf{SW} method (i.e.\textbf{RS+SW} or \textbf{DIL+SW}) can further improve the classification performance, emphasizing the significance of identifying multiple-image temporal information in personalized food classification. Finally, integrating all the modules enables the model to achieve the best performance across all methods on both benchmark datasets for most time steps.

Fig \ref{fig:results} shows the trends of classification accuracy at each time step for different methods on \textbf{Food101-Personal} dataset. In general, all methods in comparison improve over time except for \textbf{CNN}. Our proposed method shows a faster rate of improvement especially after $t=100$ as it contains more multiple-image temporal information from the past.

\vspace{-1mm}
\section{Conclusion}
In this paper, we focus on personalized food image classification. We first introduce two new benchmark datasets, Food101-Personal and VFN-Personal. Next, we propose a personalized food classifier that leverages self-supervised learning to enhance image feature extraction capabilities. We present two sampling methods, random sampling and dual instance learning, to minimize learning biases associated with sequential data, and suggest a sliding window method to capture multiple-image temporal information for the final classification. Our method is evaluated on both benchmarks and show promising improvements compared to existing work.

\vspace{-1mm}
\bibliographystyle{IEEEtran}
\small \bibliography{IEEEabrv}
\vspace{12pt}

\end{document}